# Improving probability selecting based weights for Satisfiability Problem

**Huimin Fu[1], Yang Xu[2], Jun Liu[3], Guanfeng Wu[2],** and **Sutcliffe Geoff[4]**

[1]School of Information Science and Technology, Southwest Jiaotong University, Chengdu, China
[2]National-Local Joint Engineering Laboratory of System Credibility Automatic Verification, Southwest Jiaotong University, Chengdu
[3]School of Computing, Ulster University, Northern Ireland, UK
[4]Department of Computer Science, University of Miami, USA.
Emails: fhm0807@gmail.com; xuyang@home.swjtu.edu.cn; j.liu@ulster.ac.uk; wuguanfeng@126.com; geoff@cs.miami.edu

**Abstract**

The Boolean Satisfiability problem (SAT) is important on artificial intelligence community and the impact of its solving on complex problems. Recently, great breakthroughs have been made respectively on stochastic local search (SLS) algorithms for uniform random $k$-SAT resulting in several state-of-the-art SLS algorithms Score$_2$SAT, YalSAT, ProbSAT, CScoreSAT and on a hybrid algorithm for hard random SAT (HRS) resulting in one state-of-the-art hybrid algorithm SparrowToRiss. However, there is no an algorithm which can effectively solve both uniform random $k$-SAT and HRS. In this paper, we present a new SLS algorithm named *SelectNTS* for uniform random $k$-SAT and HRS. *SelectNTS* is an improved probability selecting based local search algorithm for SAT problem. The core of *SelectNTS* relies on new clause and variable selection heuristics. The new clause selection heuristic uses a new clause weighting scheme and a biased random walk. The new variable selection heuristic uses a probability selecting strategy with the variation of CC strategy based on a new variable weighting scheme. Extensive experimental results on the well-known random benchmarks instances from the SAT Competitions in 2017 and 2018, and on randomly generated problems, show that our algorithm outperforms state-of-the-art random SAT algorithms, and our *SelectNTS* can effectively solve both uniform random $k$-SAT and HRS.

## 1 Introduction and motivation

Given a formula in clause normal form (CNF), the Boolean satisfiability (SAT) problem requires finding a Boolean assignment for the problem's variables that satisfies the formula. SAT has been widely studied as a canonical NP-complete problem. It plays a prominent role in many domains of computer science and artificial intelligence due to its significant importance in both theory and applications [1]. The SAT problem is fundamental in solving many practical problems in combinatorial optimization, statistical physics, circuit verification, computing theory [44], and SAT algorithms have been widely used to solve real-world applications, such as computer algebra systems [9], core graphs [26], scheduling [16], gene regulatory networks [17], automated verification [38], model-based diagnosis (MBD)[46], scheduling [56], machine induction [42].

Algorithms for solving SAT problems can be categorized into two main classes: complete algorithms [25, 30, 36, 37, 39, 45] and stochastic local search (SLS) algorithms [24, 33]. Although SLS algorithms are typically incomplete, they are often surprisingly effective in finding solutions to satisfiable random SAT problems [13].

In this work, we concentrate on the SLS algorithm. SLS algorithms are best suited for solving problems required short time to solve. [13]. SLS algorithms are often evaluated on random SAT instances including uniform random $k$-SAT problems [1] and hard random SAT (HRS) problems [3, 4]. The class of random SAT instances is a relatively unbiased sample for algorithms [13]. Random SAT instances remain very difficult. Indeed, such instances are challenging for all kinds of algorithms and by controlling the instance sizes and the clause-to-variable ratios, they provide adjustable hardness levels to assess the solving capabilities. Moreover, the performance of algorithm is usually stable on random SAT instances, either good or bad. Actually, the class of random SAT instances is one of the three main tracks in the well-known SAT competitions [47]. The heuristics used by SLS solvers to solve random SAT problems are also potentially useful for solving real-world SAT problems [48].

In the beginning, an SLS algorithm generally generates an initial assignment of the variables of *F*. Then it explores the search space to minimize the number of unsatisfied clauses. To do this, it iteratively flips the truth value of a variable selected according to some heuristic at each step until it seeks out a solution or timeout [27-29]. Heuristics in SLS algorithms mainly differ from each other on the variable selection heuristics at each iteration.

The representative state-of-the-art SLS algorithms include the two-mode solvers containing the configuration checking (CC) algorithms ((e.g., CCASat [14], Swqcc [50]), clause weighting algorithms (e.g., Pure Additive Weighting Scheme (PAWS) [41], Scaling and Probabilistic Smoothing (SAPS) [23] and Discrete Lagrangian Method (DLM) [43]), comprehensive score function algorithms (e.g., CScoreSAT

[13] and DCCASat [31]) and focused random walk (FRW) solvers containing the tie-breaking algorithms (e.g., WalkSATlm [12], and FrwCBlm [33]), probability selecting algorithms (e.g., ProbSAT [6,7] and YalSAT [8]), and hybrid algorithms (e.g., CCAnr [15], CSCCSat [34] Score$_2$SAT [10]) and other SAT solvers include in the literature [5, 28, 45].

Among uniform random $k$-SAT instances, random 3-SAT one exhibit some particular statistical properties and are easy to solve, for example, by SLS algorithms and a statistical physics approach called *Survey Propagation* [19]. It has been shown that the famous SLS algorithm WalkSAT [21] scales linearly with the number of variables for random 3-SAT instances near the phase transition. The state-of-the-art FrwCB solves random 3-SAT instances near the phase transition (at ratio 4.2) with millions of variables within 2-3 hours [53].

However, no single SLS heuristics can be effective on all types of random SAT instances including HRS and uniform random $k$-SAT instances with long clauses remain very difficult, since different types of instances presents different characteristics. Especially, as can be seen from the competition results of the random track of SAT Competition 2017 [54] and 2018 [55], all the participating solvers lost their power and effectiveness on several random SAT instances, especially for all HRS instances. One approach to design an effective algorithm is to balance the search between intensification and diversification to guide the algorithm. Furthermore, it is also essential to solve different types instances by developing new heuristics.

Following this spirit, we develop a new algorithm based on the basic framework of probability selecting [6], because the general SLS algorithms based on probability selecting for solving SAT include clause selection and variable selection which are two main factors affecting SLS algorithms. According the features of probability selecting algorithms, it still has two important limitations. First, like general SLS solvers, it selects a clause from unsatisfied clauses randomly. Second, using only probability selecting may result in the same variable being selected in consecutive steps. These two observations constitute the main motivations of this work. We aim to proposed an improved probability selecting based weights for random SAT, by reinforcing the original algorithm based on ProbSAT [6], which remains an innovative and appealing approach due to its selecting feature and simplicity.

We summarized the **main contribution** of this paper as follows.

- The enhanced probability selecting based on weights proposed in this work considers the feature of random SAT problem and brings two weighting schemes. First, to distinguish the unsatisfied clauses and balance the number of times each clause is selected in the clause selection, we proposed a new and global clause weighting scheme to guide the clause selection. The new clause weighting scheme, called *cNTS* (**c**lauses **N**umber of **T**imes **S**elected), that counts the number of times a clause has been selected (in Section 4.1.1). *cNTS* is different from existing clause weighting schemes that are updated according to whether or not a clause is satisfied or unsatisfied by flipping the value of a variable and only when the algorithms fall into local optimal [13, 14]. Based on *cNTS* we define *hard satisfiable clauses* (*HSCs*) (in Section 4.1.2) to distinguish unsatisfied clauses. A biased random walk guided by *cNTS* and *HSCs* is adopted as a new clause selection heuristic (in Section 4.1.3). Second, to avoid same variable being selected in consecutive steps and balance the number of times each variable is selected in the variable selection, we adopt a variation of CC strategy based on a new and global variable weighting scheme called *vNTS* (**v**ariables **N**umber of **T**imes **S**elected), that counts the number of times a variable has been selected (in Section 4.2.1). We then define a function, called $S_v$, that is a linear combination of the commonly used *score* property and *vNTS*. Variable selection uses a probability selecting method [7] with the variation of CC strategy based on $S_v$ (in Section 4.2.2). *cNTS* and *vNTS* play key roles in the two new heuristics, and thus in our new SLS algorithm, called *SelectNTS* (**Select**ion based on the **N**umber of **T**imes clauses and variables are **S**elected) (in Section 4.3).

- We assess the performance of the proposed *SelectNTS* algorithm on the benchmark instances of the well-known random track of SAT Competitions in 2017 and 2018 and generated by generators [1,3]. The experimental results show that *SelectNTS* performs remarkably well compared to state-of-the-art SLS algorithms like ProbSAT [7], YalSAT [8], Sparrow [2], CscoreSAT [13] as well as Score$_2$SAT [10] and even sophisticated hybrid algorithm called SparrowToRiss [5] on HRS instances. *SelectNTS* also proves to be competitive even when it is compared to state-of-the-art SLS algorithms like ProbSAT, YalSAT, CscoreSAT and Score$_2$SAT on uniform random $k$-SAT instances with long clauses.

This paper is structured as follows: Section 2 provides some definitions and briefly reviews some previous related heuristics. In Section 3, we review the general framework of algorithms based on probability selecting for solving random SAT. Section 4 presents our improved *SelectNTS* algorithm for random SAT. In Section 5, we illustrate some case studies. Section 6 presents the experimental results and comparisons on the random SAT instances from random track of SAT Competitions (2017 and 2018) and generated by generators. In Section 7, we summarize the main contributions of this work and suggests directions for future work.

## 2 Definitions and related heuristics for SAT

In this section, we introduce some of the basic definitions and related work to the boolean satisfiability (SAT) problem.

*Definitions*

A SAT problem $F$ in CNF is constructed from a pair $(V, C)$, where $V=\{v_1, v_2, ..., v_n\}$ is a set of $n$ Boolean variables, and $C=\{c_1, c_2, ..., c_m\}$ is a set of $m$ clauses. Each clause $c_i \in C$ is a disjunction of literals, and a literal is a variable $v_i$ or its negation. $r = m/n$ is the clause-to-variable ratio. $F$ is the conjunction of the clauses. An *assignment* for $F$ is an assignment of truth values to its variables, and a *satisfying assignment* is an assignment that makes all the clauses true.

In SLS algorithms for SAT problems, for a variable $v$ and assignment $\alpha$, $score(v, \alpha)$ is the number of increase in satisfied clauses by flipping the assigned value of $v$. and $break(v, \alpha)$ is the number of satisfied clauses that become unsatisfied by flipping the assigned value of $v$.

SLS algorithms explore the search space aiming to minimize the number of unsatisfied clauses. To do this, it is natural to select a variable in an unsatisfied clause to flip. The heuristic factors are thus the clause and variable selection.

*Related heuristics*

Although our algorithm is different from the existing SLS SAT solver, *SelectNTS* inherits some excellent features of the previous algorithm. In this section, we review briefly the existing related heuristic algorithms and the variation approach of adopting these features into our *SelectNTS*.

Heuristics in SLS algorithms for SAT can be divided into two categories: two-mode SLS algorithms and focused random walk (FRW) algorithms.

Two-mode SLS algorithms include the greedy mode and the diversification mode. In the greedy mode, to increase the number of satisfied clauses, the algorithms prefer to select the greedy variables to be flipped. In diversification mode, to avoid local optimization, the algorithms select a variable randomy to be flipped. For the two-mode in SLS SAT algorithms during the last ten decades, the most significant development was perhaps "configuration checking" strategy (CC) and "weights" strategy [18] (similar to "score function" [13]), leading to the effective CCASat [14], Swqcc [50], CScoreSAT [13] and DCCASat [31]. One of the main features of the CC strategy is that the last flipping variable must not be the current flipping variable [14] (like the simple Tabu search strategy [49]). One of the main features of the weighting schemes is that greedily select a best variable to be flipped among the candidate variables (like the well-known GSAT [51]- the score function in GSAT actually is the weight of variables). *Our SelectNTS algorithm attempts to incorporate the **ideas** of both the inhibition of CC to avoid selecting the same variable in consecutive steps and intensification of variable weighting strategy to select the best variable to be flipped among the candidate variables*.

FRW algorithms always select a variable to be flipped from an unsatisfied clause chosen randomly in each step [33]. The WalkSAT [21] is the well-known FRW algorithm. Cai et al. [12] introduced the tie-breaking strategy into the WalkSAT algorithm to prevent multiple candidate variables for solving uniform random *k*-SAT with long clauses. Cai et al. [53] improved the WalkSAT algorithm [12] by adopting the CC strategy. Luo et al. [33] adopted the tie-breaking strategy into FrwCB algorithm [53]. *Our SelectNTS algorithm borrows the **idea** of selecting an unsatisfied clause randomly to enhance its diversification capability and utilize the clause weighting strategy to distinguish unsatisfied clauses*.

The other direct improvement on WalkSAT is to extend it into a simple probability selecting strategy [20]. The ProbSAT [6] is obtained from WalkSAT by associating a probability selecting method of the variables. YalSAT [8] and polypower1.0 [52] implemented several variants of ProbSAT's algorithm. The main principle of probability selecting strategy for SAT in the literature [6-8] is that if a variable has the lowest *break* in an unsatisfied clause chosen randomly, then the variable is preferred to be selected. *We adapt this probability selecting strategy into our SelectNTS algorithm to enhance its robustness and utilize the variable weighting strategy to select the best variable to be flipped*.

In order to take advantage of both two-mode heuristics and FRW heuristics during the search process to guide the algorithm, Cai et al. [34] proposed a hybrid algorithm called CSCCSat which is a combination of DCCASat [31] and FrwCB [53], and the hybrid algorithm called Score$_2$SAT [10] is a combination of the DCCASat [31] and WalkSATlm [12]. In order to enhance the performance of SLS algorithm, Cai et al. [15] applied preprocess technology to SLS algorithms, and Balint and Manthey [5] combined the SLS algorithm and complete algorithm.

## 3 The probability selecting method for SAT [6]

The probability selecting method proposed in the literature [6] is a general framework for solving SAT problems. In this section, we briefly review the probability selecting method [6], which serves as the basis of our algorithm. The ProbSAT algorithm [6] has wide influence among current SLS algorithms and attracted increasing interest for solving SAT benchmarks in the last few years.

ProbSAT uses only the *break* values of a variable in a probability function $f(v, \alpha)$ including a polynomial or exponential shape as listed below.

$$f(v, \alpha)) = (0.9 + break(v, \alpha)))^{-cb_1} \qquad (1)$$

$$f(v, \alpha)) = (cb_2)^{-break(v, a)} \qquad (2)$$

where $cb_1$ and $cb_2$ are decimal parameters.

The ProbSAT algorithm is designed for solving SAT. The pseudo-code of ProbSAT is described in Algorithm 1 and can be found in the literature [6,7].

To apply probability selecting method to SAT problem, four processes need to be attended. The algorithm generates a complete assignment $\alpha$ randomly as the initial assignment (line 3 in Algorithm 1). During the search process, the algorithm selects an unsatisfied clause randomly (line 6 in Algorithm 1). During the probability updating process (lines 7-10 in Algorithm 1), the probability is updated by the *break* values of variables, while the probability is computed by the polynomial function in Eq. (1) for 3-SAT problems, and the probability is computed by the exponential function in Eq. (2) for the remaining problems. During the probability selecting process (line 11 in Algorithm 1), the algorithm based the probability $\frac{f(x,a)}{\sum_{z \in C} f(z,a)}$ tries to select a variable to be flipped.

ProbSAT algorithm explores the search space to minimize the number of unsatisfied clauses. To do this, it is natural for ProbSAT algorithm to select a variable to be flipped. The variable selection heuristic of ProbSAT mainly depends on

```
Algorithm 1: ProbSAT algorithm
    Input: CNF-formula F, MaxTries, MaxSteps
    Output: A satisfying assignment α of F, or "UNKNOWN"
1  begin
2      for i = 1 to MaxTries do
3          α ← a generated truth assignment randomly for F;
4          for j = 1 to MaxSteps do
5              if α satisfies F then Return α;
6              C ← an unsatisfied clause chosen at random;
7              for v in C do
8                  compute f(v, α);
9              end for
10             v ← random variable x according to probability $\frac{f(x,\alpha)}{\sum_{z \in C} f(z,\alpha)}$;
11             α ← α with v flipped;
12         end for
13     end for
14     Return "UNKNOWN";
15 end
```

two factors: clause selection strategy and variable selection strategy. In order to further improve SLS algorithms for SAT, we focus on proposing new selection heuristics, which are detailed in subsequent Sections 4.

## 4 Improving probability selecting based weights for SAT

In this section, we introduce our algorithm called *SelectNTS* based on the basic framework of probability selecting heuristic [6]. The *SelectNTS* includes two important components - clause selection heuristic based new clause weighting scheme and new variable heuristic based new variable weighting scheme.

### 4.1 The new clause selection heuristic

In this subsection we define a new clause selection heuristic, composed of three components: a clause weighting scheme called *cNTS*, a notion of *hard satisfiable clauses*, and a biased random walk.

The strategy of picking an unsatisfied clause is known to be successful for general SAT solving [6,7]. Indeed, the condition that the selected clause is unsatisfied is necessary, as selecting a satisfied clause may lead to a local optimum [51]. However, selecting from the unsatisfied clauses with equal probability does not provide enough guidance for SLS algorithms, especially for SAT problems. The number of times an unsatisfied clause is selected is an indication of how difficult it is to satisfy the clause. Based on **this observation** (in Section 5) as the basis, we propose a new clause weighting scheme to distinguish unsatisfied clauses, and in order to fully use the information of each clause in SAT problem, we propose a new clause selection heuristic to balance the number of times each clause selected

#### 4.1.1 The new and global clause weighting scheme

Clause weighting schemes such as DLM [43], SAPS [23], PAWS [41], and SWT [31] have been used successfully in SLS algorithms for general SAT solving. However, as can be seen from the competition results of the random track of SAT Competition 2017 [54] and 2018 [55], these weighting schemes are not always effective for solving different types of SAT problems. This motivated us to design a new clause weighting scheme called *cNTS*, which counts the number of times each clause has been selected.

**Definition 1** *For a clause c, in search step s, cNTS(c, s) is the number of times that c has been selected up to step s.*

- Initially, for each clause c, cNTS(c, 0)=0
- When an unsatisfied clause c is selected in step s, then
    cNTS(c, s)=cNTS(c, s-1) +1;
  Otherwise,
    cNTS(c, s)=cNTS(c, s-1).

Intuitively, clauses with larger *cNTS* values are harder to keep satisfied in the search process. Thus, it is beneficial for SLS algorithms to prefer satisfying these clauses, and we use *cNTS* to guide clause selection. The **major differences** between *cNTS* and existing clause weighting schemes are that *cNTS* is adjusted for clause, while existing clause weighting schemes are for variable, and *cNTS* is global, while existing clause weighting schemes are local [2, 11, 13, 14, 31].

#### 4.1.2 Hard satisfiable clauses

Based on *cNTS*, we define the notion of Hard Satisfiable Clauses (*HSCs*).

**Definition 2** *For a clause c, in search step s, and given a positive integer parameter β, c is an HSC in step s if and only if c is unsatisfied and cNTS(c, s) ≥ β.*

*HSCs(s,β)* denotes the set of all *HSCs* in step *s* for the given *β*. *HSCs* are regarded as good candidates for selection, especially when solving HRS problems.

#### 4.1.3 The biased random walk

An important component of most SLS algorithms is a random walk [40]. However, a standard random walk [6, 7, 8, 12] might not be suitable for HRS problems. *HSCs* are given higher priority in our algorithm by using a biased random walk [35] as follows: At each step *s*, if *HSCs* (*s, β*) is not empty an *HSC* is selected randomly, otherwise, an unsatisfied clause is selected randomly (**The biased random walk is first used in SAT problems**).

### 4.2 The new variable selection heuristic

In this subsection we define a new variable selection strategy that is composed of two components: a new variable weighting scheme called *vNTS*, and a variation of CC strategy based on a new scoring function.

In many SLS algorithms [11, 32, 35] the strategy for selecting the variable to be flipped in each step is guided by the *score* property, which maximizes the number of satisfied clauses. In contrast, we differentiate between variables in unsatisfied clauses by probability selection method, and then consider a combination of the *score* property and *the number of times each variable has been selected*. This is quite intuitive, as the more times a variable has been selected, the less likely it is that all clauses containing the variable are satisfied after subsequent variable flips. We also have used **this observation** (in Section 5) as the basis for a new variable weighting scheme, and in order to fully use the information of each variable in SAT problem, we propose a new variable selection heuristic to balance the number of times each variable selected.

*4.2.1 The new and global variable weighting scheme*

The new variable weighting scheme is defined as follows.

**Definition 3** *For a variable v, in the search step s, vNTS(v, s) is the number of times that v has been selected up to step s.*
- Initially, for each variable $v$, $vNTS(v, 0) = 0$;
- When a variable $v$ is selected in step $s$, then
$vNTS(v, s) = vNTS(v, s-1) + 1$;
Otherwise,
$vNTS(v, s) = vNTS(v, s-1)$.

Intuitively, clauses containing variables with larger *vNTS* are harder to keep satisfied in the search process, and we use *vNTS* to guide variable selection.

*4.2.2 The variation of CC strategy*

The *score* property tends to increase the number of satisfied clauses in a greedy search mode, and *vNTS* can be regarded as a heuristic for greedy search as its use tends to reduce $HSCs(s,\beta)$ by flipping the variables of an *HSC*. To combine *score* and *vNTS* in a greedy search, we define a scoring function that is a linear combination of *score* and *vNTS*, inspired by the concept of a comprehensive score [13]. The new scoring function, named $S_v$, for a variable $v$ at step $s$ when the assignment is α, is defined as follows:

**Definition 4** For a variable $v$, in search step $s$, when the assignment is $\alpha$, and given a positive integer parameter $\gamma$, $S_v(v, s, \alpha) = score(v, \alpha) + vNTS(v, s)/\gamma$.

In our algorithm, the variable is selected to be flipped firstly by the probability selecting method [6]. However, using only probability selecting method may result in the same variable being selected in consecutive steps. To avoid this, we employ a variation of CC strategy. If the variable selected by probability at step $s$ is the same as the variable flipped in step $s - 1$, a different variable with the greatest $S_v$ value is selected instead. As the variation of CC strategy, $S_v$ is very simple, and can be computed with little overhead.

This variation of CC strategy is inspired by the idea in the literature [14], but is essentially different. In our algorithm a variable is selected to be flipped from an unsatisfied clause selected randomly, and there is no need to select candidate variables from all variables.

*4.3 The SelectNTS Algorithm*

Based on the above ideas, this section presents *SelectNTS*, whose pseudo code is shown in Algorithm 2.

*SelectNTS* has an outer loop that (re)starts with a randomly generated truth assignment, looping maximally *MaxTries* times (lines 1-2). *bestVar* is used to record which variable was flipped in the last step (line 3). Within that outer loop the inner loop searches for a satisfying assignment with up to *MaxSteps* variable flips (line 4). If the current assignment satisfies all clauses of *F* then *SelectNTS* returns the assignment (line 5). Otherwise *SelectNTS* proceeds with the biased random walk: if $HSCs(s, \beta)$ is not empty (line 6), an *HSC* is selected randomly (line 7), otherwise, an unsatisfied clause is selected randomly (line 8). *cNTS* is updated for the

**Algorithm 2** The *SelectNTS* Algorithm

**Input**: CNF-formula *F*, *MaxTries*, *MaxSteps*, γ, β
**Output**: A satisfying assignment α of *F*, or *Unknown*
1:  **for** *try:* = 1 to *MaxTries* **do**
2:      α := a randomly generated truth assignment;
3:      *bestVar* := null;
4:      **for** *step:*= 1 to *MaxSteps* **do**
5:          **if** α satisfies *F* **then return** α;
6:          **if** $HSCs(step,\beta) \neq \emptyset$ **then**
7:              *C* := a random *HSC*;
8:          **else** *C* := a random unsatisfied clause;
9:          update *cNTS*;
10:         $v := x \in C$ selected with probability $\frac{f(x,\alpha)}{\sum_{z \in C} f(z,\alpha)}$;
11:         **if** $v :== bestVar$ **then**
12:             *bestVar* := $x \in C$, $x \neq v$, with greatest $S_v(x,s,\alpha)$;
13:         **else** *bestVar* := *v*;
14:         update *vNTS*;
15:         α := α with *bestVar* flipped;
16: **return** *Unknown*;

selected clause (line 9). *SelectNTS* then firstly picks a variable from the selected clause by the probability selecting method (line10), and if the variable is the same as *bestVar* (line 11) then *SelectNTS* selects a variable with the greatest $S_v$ value (line 12). *vNTS* is updated for the selected variable (line 14). *SelectNTS* then flips the assignment value of the chosen variable (line 15) and starts the next search step. If *MaxTries* is reached *SelectNTS* reports *Unknown* (line 16).

## 5 Case study

In Section 4.1 and Section 4.2, we introduce the definitions of *cNTS* and *vNTS* respectively, and then based on the observation of distributions of *cNTS* and *vNTS* within a certain step, we propose the new clause selection heuristic and variable selection heuristic respectively.

Different types of SAT problems may provide different distributions of *cNTS* and *vNTS* respectively. To do so, we have studied numerous SAT instances with the aim of determining the distribution of each of them. In this section, we present two case studies (at HRS and uniform random *k*-SAT from SAT Competition 2017 [54]) about distributions of *cNTS* and *vNTS* within $10^5$ steps in both the original probability selecting algorithm [7] and our *SelectNTS* algorithm (parameter settings in Section 6.2) respectively.

*5.1 Case 1: fla-qhid-540-5*

In this first case, we study a HRS instance which distributions correspond to the *cNTS* (in Section 4.1.1) and *vNTS* (in Section 4.2.1) within $10^5$ steps in both ProbSAT and our algorithm *SelectNTS* respectively. Table 1 exhibits several important information of the HRS instance fla-qhid-540-5 from random track of SAT Competition 2017.

Fig.1 and Fig.2 illustrate the distributions of *cNTS* and *vNTS* for the instance within $10^5$ steps on ProbSAT. Fig.3 and Fig.4 illustrate the distributions of *cNTS* and *vNTS* for the instance within $10^5$ steps on *SelectNTS*.

**Table 1**
Statistical description of HRS instances fla-qhid-540-5.

| Benchmark's name | fla-qhid-540-5 |
|---|---|
| Number of clauses | 2970 |
| Number of variables | 540 |
| ratio | 5.5 |
| Generator seed | 5 |

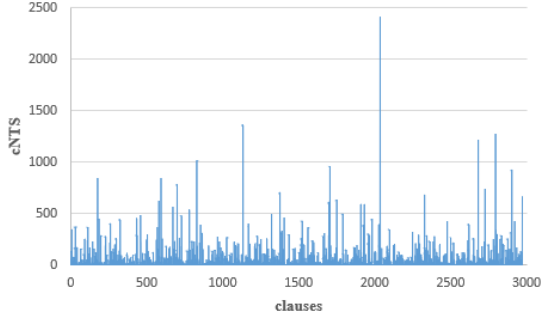

**Fig.1.** fla-qhid-540-5 Distribution of *cNTS* within $10^5$ steps on ProbSAT.

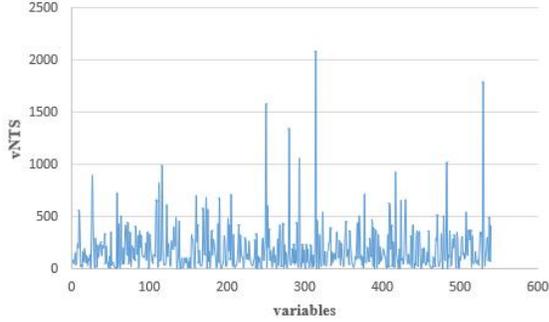

**Fig.2.** fla-qhid-540-5 Distribution of *vNTS* within $10^5$ steps on ProbSAT.

According to Fig.1, the number of times each clause is selected is quite different. Within $10^5$ steps, the maximum value of *cNTS* is close to 2500. From Fig.2, the distribution of *vNTS* is also very uneven, and the maximum value of *vNTS* is close to 2000 within $10^5$ steps.

The intuition is that the larger *cNTS* of a clause is an indication of how difficult it is to keep the clause satisfied, and as the larger *vNTS* of a variable, the less likely it is that all clauses containing the variable are satisfied after subsequent variable flips. In order to fully use the information of each clause and variable in SAT problem, we propose the new clause selection heuristic (in Section 4.1) and new variable selection heuristic (in Section 4.2) to balance the number of times each clause selected and each variable selected respectively. Finally, the distributions of *cNTS* and *vNTS* for the instance within $10^5$ steps on *SelectNTS* are presented in Fig.3 and Fig.4 respectively.

Comparing Fig. 3 with Fig. 1, and Fig. 4 with Fig. 2, the distributions of *cNTS* and *vNTS* are relatively more **uniform** on *SelectNTS* than of that on ProbSAT respectively. Specially, the maximum value of *cNTS* on *SelectNTS* which is about **0.4** times as large as on ProbSAT, is close to 1000. The maximum value of *vNTS* on *SelectNTS* which is about **0.5** times as large as on ProbSAT, is close to 1000. It follows that the new clause selection heuristic and variable selection heuristic dramatically balances the values of *cNTS* and *vNTS* for this HRS instances respectively. Thus, the new clause selection heuristic and the new variable selection heuristic play the important roles in *SelectNTS*.

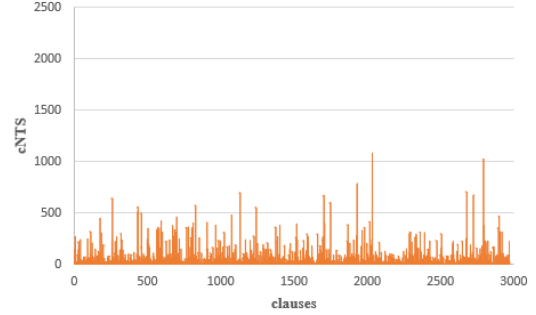

**Fig.3.** fla-qhid-540-5 Distribution of *cNTS* within $10^5$ steps on *SelectNTS*.

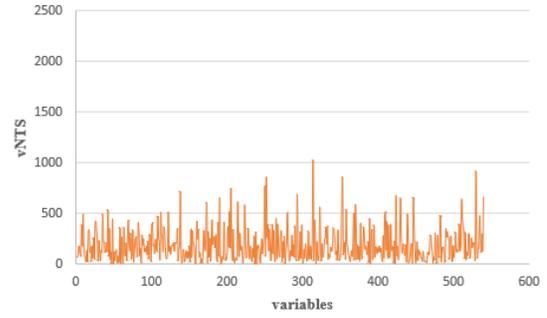

**Fig.4.** fla-qhid-540-5 Distribution of *vNTS* within $10^5$ steps on *SelectNTS*.

*5.2 Case 2: unif-k5-r21.117-v540-c11403*

In this second case, we study a uniform random *k*-SAT instance which distributions corresponds to the *cNTS* and *vNTS* within $10^5$ steps in both ProbSAT and *SelectNTS* respectively. Table 2 summarizes the most important information of the *unif-k5-r21.117-v540-c11403* instance from the random track of SAT Competition 2017.

Fig.5 and Fig.6 illustrate the distributions of *cNTS* and *vNTS* for the instance within $10^5$ steps on ProbSAT. Fig.7 and Fig.8 illustrate the distributions of *cNTS* and *vNTS* for the instance within $10^5$ steps on *SelectNTS*.

**Table 2**
Statistical description of the instances unif-k5-r21.117-v540-c11403.

| Benchmark's name | unif-k5-r21.117-v540-c11403 |
|---|---|
| Number of clauses | 11403 |
| Number of variables | 540 |
| ratio | 21.117 |
| Generator seed | 5955214796121725857 |

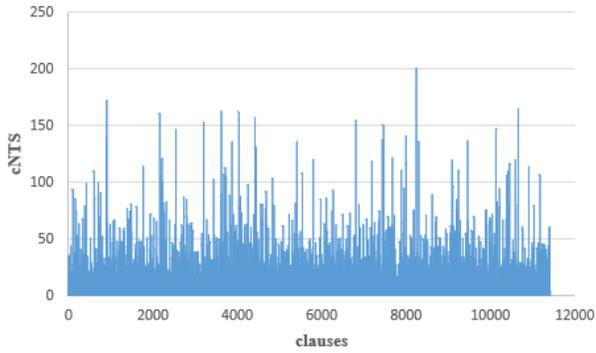

**Fig.5.** unif-k5-r21.117-v540-c11403 Distribution of *cNTS* within $10^5$ steps on ProbSAT.

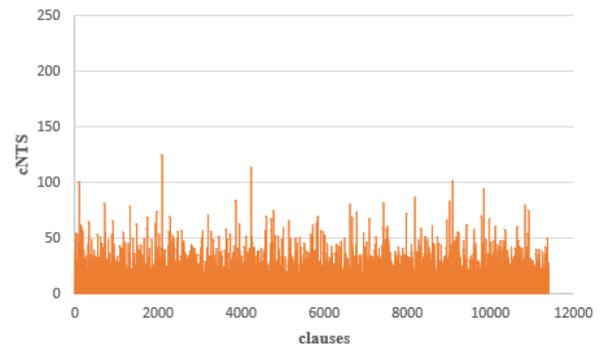

**Fig.7.** unif-k5-r21.117-v540-c11403 Distribution of *cNTS* within $10^5$ steps on *SelectNTS*.

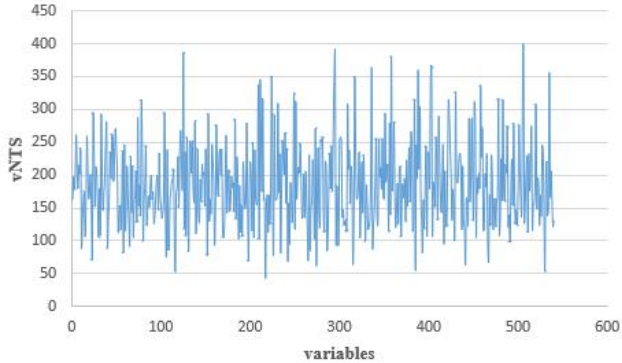

**Fig.6.** unif-k5-r21.117-v540-c11403 Distribution of *vNTS* within $10^5$ steps on ProbSAT.

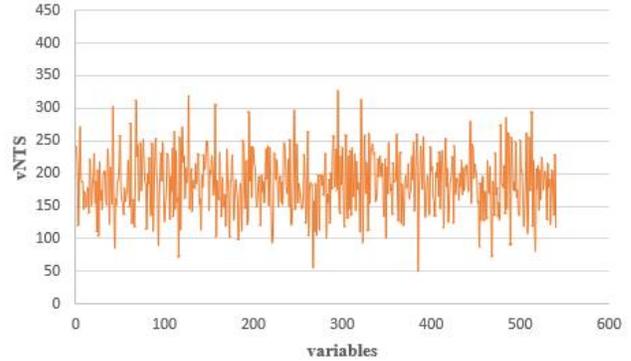

**Fig.8.** unif-k5-r21.117-v540-c11403 Distribution of *vNTS* within $10^5$ steps on *SelectNTS*.

Comparing Fig. 5 with Fig. 7, and Fig. 6 with Fig. 8, the distributions of *cNTS* and *vNTS* are relatively more **uniform** on *SelectNTS* than of that on ProbSAT respectively. Specially, the maximum value of *cNTS* is about 200 on ProbSAT from Fig.5, while the maximum value of *cNTS* is close to 150 on *SelectNTS* from Fig.7, i.e., the maximum value of *cNTS* on *SelectNTS* is about **0.6** times as large as on ProbSAT. The maximum value of *vNTS* is about 400 on ProbSAT from Fig.6, while the maximum value of *cNTS* is close to 350 on *SelectNTS* from Fig.8, i.e., the maximum value of *vNTS* on *SelectNTS* is about **0.8** times as large as on ProbSAT. It follows that our algorithm *SelectNTS* dramatically can also balance the values of *cNTS* and *vNTS* for this uniform random SAT instances respectively.

Comparing Fig. 1 with Fig. 3, Fig. 2 with Fig. 4, Fig. 5 with Fig. 7 and Fig. 6 with Fig. 8, the distributions of *cNTS* and *vNTS* on the HRS instances changes more obviously than of that on the uniform random SAT instance respectively. Thus, comparing case 1 with case 2, the new clause selection heuristic and variable selection heuristic on distributions of *cNTS* and *vNTS* for the HRS instance have more influence than of that for the uniform random SAT instance under the same variable size. We conjecture that the performance of *SelectNTS* in solving different random SAT problems may be related to degree of influencing distributions of *cNTS* and *vNTS* on *SelectNTS* compared to ProbSAT.

## 6 Experimental evaluations

We carried out experiments to evaluate *SelectNTS* on random SAT problems. For each class, we compared *SelectNTS* with state-of-the-art SLS solvers and a hybrid solver.

### 6.1 Benchamarks

All the HRS problems used in our experiments were generated by the HRS tool [3], and all the uniform random k-SAT problems utilized were generated by generator [56]. Specifically, we used the following problems (where *r* is the clause-to-variable ratio, and *n* is the number of variables).

- **SAT Competition 2017:** SAT problems taken from the random track of the SAT Competition 2017. All random HRS instances (120 instances, 40 for each ratio *r*, *r*=4.3, 5.206, 5.5, 15 instances for each variable *n* = 400, 420, 440, …, 540), which vary in both size and ratio. All uniform random *k*-SAT instances with *k*>3 (120 instances, 60 for each *k*-SAT, *k*=5, 7), which vary in both size and ratio. The uniform random 5-SAT instances vary from 200 variables at the threshold ratio of phase transition *r*=21.117 to 590 variables, from 16.0 ratio at *n*=250000 to 19.8 ratio. The uniform random 7-SAT instances vary from 90 variables at the threshold ratio of phase transition *r*=87.79 to 168 variables, from 55.0 ratio at *n*=50000 to 74.0 ratio. These HRS and uniform random instances occupy 80% of the random benchmark in SAT Competition 2017, indicating that the importance of these instances.

- **HRS Random 5.206**: HRS problems generated by the HRS tool. $r = 5.206$, $n = 600, 700, 800, 900$, and $1000$ (1000 instances, 200 instances for each size).
- **HRS Random 5.5**: HRS problems generated by the HRS tool. $r = 5.5$, $n = 600, 700, 800, 900$, and $1000$ (1000 instances, 200 instances for each size).
- **HRS Random 5.699**: HRS problems generated by the HRS tool. $r = 5.699$, $n = 200, 300, 400, 500, 600, 700, 800, 900$, and $1000$ (900 instances, 100 instances for each size).
- **HRS Random 7.821**: HRS problems generated by the HRS tool. $r = 7.821$, $n = 200, 300, 400, 500, 600, 700, 800, 900$, and $1000$ (900 instances, 100 instances for each size).
- **Uniform random 5-SAT:** Random 5-SAT problems generated by the $k$-SAT generator [56] (250 instances). *Medium 5-SAT instances* at the threshold ratio of phase transition ($r=21.115$, 100 instances, $n=200, 250, 300, 350, 400$, 20 instances for each size). *Huge 5-SAT instances* at $r<21.117$ ($n=250000$, 150 instances, $r=18.0$, $r=18.2$, $r=18.4$, 50 instances for each ratio).
- **Uniform random 7-SAT:** Random 7-SAT problems generated by the $k$-SAT generator (250 instances). *Medium 7-SAT instances* at the threshold ratio of phase transition ($r=87.79$, 100 instances, $n=100, 110, 120, 130, 140$ 20 instances for each size). *Huge 7-SAT instances* at $r< 87.79$ ($n=50000$, 150 instances, $r=65.0$, $r=66.0$, $r=67.0$, 50 instances for each ratio).
- **SAT Competition 2018**: the benchmark from the random track of SAT Competition 2018. The HRS instances and uniform random $k$-SAT instances with $k>3$ have various sizes and ratios. These instances 88.2% of the random benchmark in SAT Competition 2018.

### 6.2 Experimental Preliminaries

**Implementation**: *SelectNTS* is implemented in C. We tuned the $\beta$ and $\gamma$ parameters of *SelectNTS* according to our experience in Table 3 and Table 4. For $cb_1$ and $cb_2$, we utilize the default parameter setting tuned in the literature [7].

**Table 3**
Parameter settings of *SelectNTS* for HRS instances.

|  | $r=4.3$ | $r=5.206$ | $r=5.5/5.699$ | $r=7.821$ |
|---|---|---|---|---|
| $n\leq600$ |  | $\beta = 80$ | $\beta = 110$ |  |
|  | $\beta = 10$ | $\gamma =300$ | $\gamma =1200$ | $\beta = 400$ |
| $n>600$ | $\gamma =1200$ | $\beta = 60$ | $\beta = 110$ | $\gamma =300$ |
|  |  | $\gamma =800$ | $\gamma =900$ |  |

**Table 4**
Parameter settings of *SelectNTS* for uniform random $k$-SAT.

|  | 5-SAT | 7-SAT |
|---|---|---|
| medium instances | $\beta = 5000000$ | $\beta = 700000$ |
|  | $\gamma =500000$ | $\gamma =500000$ |
| huge instances | $\beta = 700$ | $\beta = 2000$ |
|  | $\gamma =600$ | $\gamma =4000$ |

**Competitors:** In the following we use RSC to denote **R**andom track of the **SAT C**ompetition. In order to evaluate the relative effectiveness and efficiency of *SelectNTS*, we compare *SelectNTS* with the following a hybrid solver and 4 state-of-the-art SLS SAT solvers:

- SparrowToRiss (denoted by STR in the result tables) [5]: a hybrid algorithm is the 1st place in RSC 2018.
- ProbSAT [7]: The probability selecting algorithm is the 2nd place among the SLS algorithms in RSC 2018 and 1st place in RSC 2013.
- YalSAT [8]: The probability selecting algorithm is the 1st place in RSC 2017.
- Score$_2$SAT [10]: The algorithm with CC strategy and clause weighting scheme is the 2nd place in RSC 2016.
- CScoreSAT [13]: The algorithm with CC strategy, clause weighting scheme and complex scoring functions.

ProbSAT and SparrowToRiss, are obtained from the web site of the SAT Competition 2018 [55]. YalSAT and Score$_2$SAT is obtained from the web site of the SAT Competition 2017 [54]. CScoreSAT is obtained from the web site of SAT Competition 2013 [57].

**Evaluation Methodology**: The experiments are carried out on an Intel(R) Core (TM) i7-6700M 3.4 GHz CPU with 16GB RAM, running the 64-bit Ubuntu Linux operating system. The CPU time limit is 600 seconds for the HRS Random 5.206, 5.5, 5.699, 7.821 problem sets (as in the literature [4]), and 5000 seconds for remaining benchmarks (as in the SAT competitions in 2017 and 2018).

For all benchmarks, we run each solver 10 times for each instance. For performance metrics, we report the number of average solved instances at ten run "AverS", and the penalized average run time "PAR-2" (an unsuccessful run is penalized two times the time limit) as in the competitions. Note that for cases of 100% success rate PAR-2 is the average CPU time taken. The best results for an instance class are highlighted in **bold**.

### 6.3 Experimental Results

In this subsection, we present the comparative experimental results of *SelectNTS* and its competitors on each problem set.

**Results for SAT Competition 2017**

Table 5 presents the results of the performance of *SelectNTS* compared with 4 state of the art SLS solvers and a hybrid solver on all HRS and uniform random $k$-SAT with long clauses from SAT Competition 2017.

Since *SelectNTS* is based on ProbSAT, we first compare these two solvers. From Table 5, although ProbSAT spends a few less time than *SelectNTS* for the HRS instances with $r=4.3$ and the uniform 7-SAT instances with $r=87.79$, ProbSAT and *SelectNTS* solve the same number of instances. For remaining instance classes, *SelectNTS* solves more instances than ProbSAT. Overall, ProbSAT solves 102 instances on average, while *SelectNTS* solves 177 instances on average, which is 1.74 times as many as ProbSAT does.

*SelectNTS* solves more instances than its competitors. Overall, *SelectNTS* solves 177 instances on average, compared to 158 for SparrowToRiss on average and 102 for Score$_2$SAT on average and 100 for both YalSAT and CScoreSAT on average. Further observation shows that, for HRS instances classes, *SelectNTS* significantly outperforms SparrowToRiss which is the current best solver for HRS instances [55], and for solving uniform random *k*-SAT instances with *k*>3, *SelectNTS* significantly outperforms state-of-the-art SLS solvers - CScoreSAT, YalSAT, Score$_2$SAT and ProbSAT which are among the most successful solvers for solving uniform random *k*-SAT instances in the literature.

**Table 5**
Computational results on the **SAT Competition 2017** benchmark.

| Random SAT | Ratio | STR | | CScoreSAT | | Score$_2$SAT | | YalSAT | | PobSAT | | *SelectNTS* | |
|---|---|---|---|---|---|---|---|---|---|---|---|---|---|
| | | AverS | PAR-2 | AverS | PAR-2 | AverS | PAR-2 | AverS | PAR-2 | AverS | PAR-2 | AverS | PAR-2 |
| HRS | 4.3 | 40 | 0.117 | 40 | **0.009** | 40 | 0.008 | 40 | 0.017 | 40 | 0.057 | 40 | 0.134 |
| | 5.206 | 40 | 5.709 | 0 | - | 0 | - | 0 | - | 0 | - | 40 | **0.032** |
| | 5.5 | 40 | 151.0 | 6 | 8500 | 9 | 7750 | 9 | 7750 | 9 | 7750 | 40 | **0.113** |
| Uniform | <21.117 | 4 | 8083 | 10 | 5250 | 8 | 6231 | 12 | 4147 | 11 | 4526 | **13** | **3831** |
| | 21.117 | 9 | 7760 | 15 | **6476** | 14 | 6655 | 13 | 6880 | 13 | 6829 | 14 | 6661 |
| | <87.79 | 9 | 5602 | 11 | 4839 | 11 | 5756 | 9 | 5517 | 11 | 4514 | **12** | **4045** |
| | 87.79 | 16 | 6035 | 18 | 5931 | **19** | **5582** | 17 | 5957 | 18 | 5552 | 18 | 5800 |
| Overall/240 | | 158 | 3466 | 100 | 5992 | 101 | 5997 | 100 | 5903 | 102 | 5775 | **177** | **2733** |

**Table 6**
Computational results on the **HRS Random 5.206** benchmark.

| | n=600 AverS PAR-2 | n=700 AverS PAR-2 | n=800 AverS PAR-2 | n=900 AverS PAR-2 | n=1000 AverS PAR-2 |
|---|---|---|---|---|---|
| STR | 200 11.25 | 200 11.81 | 160 273.0 | 120 505.2 | 120 504.5 |
| ProbSAT | 0 - | 0 - | 0 - | 0 - | 0 - |
| YalSAT | 0 - | 0 - | 0 - | 0 - | 0 - |
| Score$_2$SAT | 0 - | 0 - | 0 - | 0 - | 0 - |
| CScoreSAT | 0 - | 0 - | 0 - | 0 - | 0 - |
| *SelectNTS* | 200 **0.038** | 200 **0.073** | **200** **0.088** | **200** **0.065** | **200** **0.145** |

**Table 7**
Computational results on the **HRS Random 5.5** benchmark.

| | n=600 AverS PAR-2 | n=700 AverS PAR-2 | n=800 AverS PAR-2 | n=900 AverS PAR-2 | n=1000 AverS PAR-2 |
|---|---|---|---|---|---|
| STR | 200 5.812 | 140 363.6 | 40 960.4 | 80 722.2 | 0 - |
| ProbSAT | 40 960.0 | 40 960.0 | 40 960.0 | 40 960.0 | 0 - |
| YalSAT | 40 960.0 | 40 960.0 | 40 960.0 | 40 960.0 | 0 - |
| Score$_2$SAT | 40 960.0 | 40 960.0 | 40 960.0 | 0 - | 0 - |
| CScoreSAT | 0 - | 0 - | 0 - | 0 - | 0 - |
| *SelectNTS* | 200 **0.179** | **200** **0.217** | **200** **0.258** | **200** **0.339** | **200** **0.314** |

**Table 8**
Computational results on the **HRS Random 5.699** benchmark.

| Variables | STR | | CScoreSAT | | Score$_2$SAT | | YalSAT | | PobSAT | | *SelectNTS* | |
|---|---|---|---|---|---|---|---|---|---|---|---|---|
| | AverS | PAR-2 | AverS | PAR-2 | AverS | PAR-2 | AverS | PAR-2 | AverS | PAR-2 | AverS | PAR-2 |
| n=200 | 100 | 42.01 | 0 | - | 0 | - | 0 | - | 0 | - | 100 | **0.023** |
| n=300 | 100 | 94.28 | 0 | - | 0 | - | 0 | - | 0 | - | 100 | **0.037** |
| n=400 | 100 | 213.5 | 0 | - | 0 | - | 0 | - | 0 | - | 100 | **0.067** |
| n=500 | 100 | 228.2 | 0 | - | 0 | - | 0 | - | 0 | - | 100 | **0.080** |
| n=600 | 80 | 444.1 | 0 | - | 0 | - | 0 | - | 0 | - | **100** | **0.102** |
| n=700 | 40 | 855.9 | 0 | - | 0 | - | 0 | - | 0 | - | **100** | **0.128** |
| n=800 | 0 | - | 0 | - | 0 | - | 0 | - | 0 | - | **100** | **0.157** |
| n=900 | 0 | - | 0 | - | 0 | - | 0 | - | 0 | - | **100** | **0.185** |
| n=1000 | 0 | - | 0 | - | 0 | - | 0 | - | 0 | - | **100** | **0.197** |

**Table 9**
Computational results on the **HRS Random 7.821** benchmark.

| Variables | STR | | CScoreSAT | | Score$_2$SAT | | YalSAT | | PobSAT | | *SelectNTS* | |
|---|---|---|---|---|---|---|---|---|---|---|---|---|
| | AverS | PAR-2 | AverS | PAR-2 | AverS | PAR-2 | AverS | PAR-2 | AverS | PAR-2 | AverS | PAR-2 |
| n=200 | 100 | 202.6 | 0 | - | 0 | - | 0 | - | 0 | - | 100 | **0.224** |
| n=300 | 100 | 213.8 | 0 | - | 0 | - | 0 | - | 0 | - | 100 | **0.423** |
| n=400 | 100 | 219.1 | 0 | - | 0 | - | 0 | - | 0 | - | 100 | **0.657** |
| n=500 | 100 | 236.6 | 0 | - | 0 | - | 0 | - | 0 | - | 100 | **0.874** |
| n=600 | 100 | 252.9 | 0 | - | 0 | - | 0 | - | 0 | - | 100 | **1.029** |
| n=700 | 100 | 253.9 | 0 | - | 0 | - | 0 | - | 0 | - | 100 | **1.468** |
| n=800 | 100 | 279.8 | 0 | - | 0 | - | 0 | - | 0 | - | 100 | **1.550** |
| n=900 | 100 | 277.3 | 0 | - | 0 | - | 0 | - | 0 | - | 100 | **1.911** |
| n=1000 | 100 | 314.9 | 0 | - | 0 | - | 0 | - | 0 | - | 100 | **2.271** |

The sizes of HRS instances from SAT Competition 2017 are not large enough to provide a good spectrum of instances for solvers. In order to investigate the detailed performance of *SelectNTS* and solvers on HRS instances with *r*=5.206, 5.5, 5.699 and 7.821, we evaluate them on HRS Random benchmark, where the instance size increases more quickly.

### Results for HRS Random 5.206

Table 6 suggests that the difficulty of such HRS instances with *r*=5.206 increases significantly with a relatively large increment of the size. According to Table 6, the results show *SelectNTS* dramatically outperforms its competitors. For example, all SLS competitors fail in all instance classes, and SparrowToRiss solves 200, 200, 160, 120, 120 instances for each instance class on average respectively, while *SelectNTS* solves all instances for each instance class on average. Indeed, to be best of our knowledge, all HRS Random 5.206 instances are solved for the first time. Given the good performance of *SelectNTS* on HRS Random 5.206 instances with 1000 variables, it is very likely it could be able to solve larger HRS instances with *r*=5.206.

### Results for HRS Random 5.5

To measure the performance of *SelectNTS* on HRS instances with *r*=5.5 more accurately, we additionally test *SelectNTS* on the HRS Randm 5.5 benchmark generated by the HRS tool, compared with Score$_2$SAT, CScoreSAT, YalSAT, ProbSAT, and SparrowToRiss.

The results are presented in Table 7. It is clear that *SelectNTS* shows significantly better performance than all its competitors on the whole benchmark. Also, *SelectNTS* outperforms its competitors in terms of par 2, which is more obvious as the instance size increases. In particular, on the HRS instance with *n*=900, Score$_2$SAT and CScoreSAT fail in all instances, and the other competitors succeed in less than 80 instances on average, while *SelectNTS* solves all instances (200) on average. Finally, *SelectNTS* is the only solver that solves all HRS Random 5.5 benchmark on average, which illustrates its robustness.

### Results for HRS Random 5.699

We conduct more empirical evaluations of *SelectNTS* with its SLS solvers and a hybrid solver on HRS instances with *r*=5.699, the benchmark is generated by HRS tool [3].

The experimental results are presented in Table 8. For HRS instances with *n*=200, 300, 400, 500 classes, *SelectNTS* and SparrowToRiss solve the same number of instances on average, but *SelectNTS* has less accumulative run time. For HRS instances with *n*=600, 700, 800, 900 and 1000 classes, *SelectNTS* solves the most instances. Especially, *SelectNTS* shows significantly superior performance than its competitors on HRS instances with *n*=800, 900, 1000 classes, where it solves 100 instances on average for each class, while its competitors fail to find a solution for any of these instance classes.

### Results for HRS Random 7.821

We conduct more empirical evaluations of *SelectNTS* with Score$_2$SAT, CScoreSAT, YalSAT, ProbSAT, and SparrowToRiss on HRS instances with *r*=7.821.

Table 9 presents the experimental results of *SelectNTS* and its competitors on the HRS Random 7.821 benchmark. It is promising to see the performance of *SelectNTS* remains surprisingly good on these HRS random 7.821 benchmark, where its competitors show rather poor performance, especially for SLS solvers. For the experimental results, Score$_2$SAT, CScoreSAT, YalSAT, ProbSAT fail to find a solution for the whole benchmark, while *SelectNTS* solves all instances on average. Although *SelectNTS* and SparrowToRiss solve the same number of instances on average for the whole benchmark, but *SelectNTS* is over 138 times faster than SparrowToRiss on average in the whole HRS Random 7.821 instances indicating that *SelectNTS* is the comprehensive best algorithm in this comparison. On the other hand, SparrowToRiss is the first place on the random SAT track of SAT Competition 2018, thus it is challenging to improve such performance over SparrowToRiss, indicating that *SelectNTS* algorithm achieves the state-of-the-art performance on HRS instances with *r*=7.821.

**Table 10**
Computational results on the **Uniform random 5-SAT** benchmark.

| Ratio | Variable | STR | | CScoreSAT | | Score$_2$SAT | | YalSAT | | PobSAT | | *SelectNTS* | |
|---|---|---|---|---|---|---|---|---|---|---|---|---|---|
| | | AverS | PAR-2 | AverS | PAR-2 | AverS | PAR-2 | AverS | PAR-2 | AverS | PAR-2 | AverS | PAR-2 |
| Medium instances | | | | | | | | | | | | | |
| | *n*=200 | 11 | 4516 | 11 | 4506 | 11 | 4523 | 11 | 4513 | 11 | 4513 | 11 | **4501** |
| | *n*=250 | 9 | 5582 | 10 | 5069 | 9 | 5502 | 10 | 5247 | 10 | 5142 | 10 | **5022** |
| *r*=21.117 | *n*=300 | 3 | 8525 | 8 | 6298 | 9 | 6078 | 10 | **5283** | 8 | 6122 | 9 | 5993 |
| | *n*=350 | 8 | 6091 | 12 | 4166 | 13 | 3749 | 13 | 3734 | 13 | 3734 | 13 | **3703** |
| | *n*=400 | 1 | 9510 | 3 | 8667 | 3 | 8728 | 2 | 9216 | 3 | **8613** | 3 | 8703 |
| Huge instances | | | | | | | | | | | | | |
| *r*=18.0 | | 0 | - | 0 | - | 50 | 3768 | 50 | 354.2 | 48 | 831.0 | 50 | **476.5** |
| *r*=18.2 | *n*=2.5×10$^5$ | 0 | - | 0 | - | 0 | - | 46 | 1626 | 0 | - | **50** | **1558** |
| *r*=18.4 | | 0 | - | 0 | - | 0 | - | 0 | - | 0 | - | **50** | **3900** |

**Table 11**
Computational results on the **Uniform random 7-SAT** benchmark.

| Ratio | Variable | STR | | CScoreSAT | | Score$_2$SAT | | YalSAT | | PobSAT | | *SelectNTS* | |
|---|---|---|---|---|---|---|---|---|---|---|---|---|---|
| | | AverS | PAR-2 | AverS | PAR-2 | AverS | PAR-2 | AverS | PAR-2 | AverS | PAR-2 | AverS | PAR-2 |
| Medium instances | | | | | | | | | | | | | |
| | $n$=100 | 12 | **4013** | 12 | 4043 | 12 | 4037 | 12 | 4044 | 12 | 4044 | 12 | 4042 |
| | $n$=110 | 10 | 5087 | 10 | 5141 | 11 | 4592 | 11 | 4749 | 11 | 4559 | 11 | **4558** |
| $r$=87.79 | $n$=120 | 9 | 5626 | 9 | 5780 | 10 | 5248 | 10 | 5451 | 9 | 5969 | **11** | 5442 |
| | $n$=130 | 10 | 5123 | 10 | 5518 | 13 | **3981** | 13 | 4412 | 12 | 4380 | 12 | 4324 |
| | $n$=140 | 10 | 5087 | 11 | 4829 | 13 | **4048** | 10 | 5397 | 10 | 5597 | 13 | 4566 |
| Huge instances | | | | | | | | | | | | | |
| $r$=65.0 | | 0 | - | 40 | 4860 | 41 | 5234 | 0 | - | 47 | 778.2 | **50** | **268.2** |
| $r$=66.0 | $n$=5×10$^4$ | 0 | - | 0 | - | 0 | - | 0 | - | 0 | - | **50** | **1444** |
| $r$=67.0 | | 0 | - | 0 | - | 0 | - | 0 | - | 0 | - | **18** | **7484** |

**Table 12**
Computational results on the **SAT Competition 2018** benchmark.

| Random SAT | Ratio | STR | | CScoreSAT | | Score$_2$SAT | | YalSAT | | PobSAT | | *SelectNTS* | |
|---|---|---|---|---|---|---|---|---|---|---|---|---|---|
| | | AverS | PAR-2 | AverS | PAR-2 | AverS | PAR-2 | AverS | PAR-2 | AverS | PAR-2 | AverS | PAR-2 |
| | 4.3 | 55 | 0.052 | 55 | 0.009 | 55 | **0.001** | 55 | **0.001** | 55 | 0.013 | 55 | 0.032 |
| HRS | 5.206 | 55 | 1.020 | 8 | 8591 | 33 | 4000 | 9 | 8387 | 12 | 7858 | 55 | **0.017** |
| | 5.5 | 55 | 136.4 | 11 | 8000 | 12 | 7818 | 12 | 7818 | 12 | 7818 | 55 | **0.060** |
| | <21.117 | 3 | 8570 | 9 | 5706 | 11 | 4683 | 12 | 4079 | 11 | 4524 | **13** | **3821** |
| Uniform | 21.117 | 7 | 3111 | 8 | 2495 | 7 | 3015 | 8 | 2326 | 7 | 3404 | **9** | **2015** |
| | <87.79 | 9 | 5657 | 10 | 5129 | 11 | 4720 | 9 | 5520 | 11 | 4522 | **14** | **3748** |
| | 87.79 | 8 | 2262 | 5 | 5224 | 8 | 2453 | 6 | 4488 | 8 | 2967 | **9** | **1696** |
| Overall/225 | | 192 | 1537 | 106 | 5362 | 137 | 3968 | 111 | 5117 | 116 | 4919 | **209** | **837.8** |

The sizes of uniform random instances from SAT Competition 2017 are not enough to provide a good spectrum of instances for solvers. In order to investigate the detailed performance of *SelectNTS* and state-of-the-art solvers on uniform medium and huge random $k$-SAT instances with long clauses, we evaluate them on uniform Random $k$-SAT benchmarks, where the instance size increases more quickly on medium uniform random $k$-SAT with $k$>3.

**Results for Uniform Random 5-SAT**

We present in Table 10 the experimental results of *SelectNTS* for uniform medium and huge 5-SAT instances. Table 10 indicates our *SelectNTS* algorithm performs quite well on this uniform random 5-SAT instances. Specially, for all 7 instance classes, *SelectNTS* shows the best performance for 5 instance classes, while YalSAT and PrboSAT show the best performance only for one instance class and the remaining three algorithms SparrowToRiss, CScoreSAT and Score$_2$SAT cannot show the best performance for any instance classes.

Especially, for medium 5-SAT instances with $n$=200, 250, 350, *SelectNTS*, YalSAT and PrboSAT solve the same number of instances on average, but *SelectNTS* has less run time. For medium 5-SAT instances with $n$=300, *SelectNTS* has similar performance as the best solver YalSAT, solving only one less instance on average. For medium 5-SAT instances with $n$=400, although ProbSAT has less run time, *SelectNTS* and ProbSAT solve the same number of instances on average. The huge 5-SAT instances with a few million clauses and the ratio from far from the phase-transition ratio to relatively close, are as large as some of the application benchmarks. As can be seen form Table 10, *SelectNTS* is based on ProbSAT, while *SelectNTS* solves more instances than ProbSAT. Overall, ProbSAT solves 48 (out of 150) instances on average, while *SelectNTS* solves all instances, which is 3 times as many as ProbSAT does. *SelectNTS* solves more instances than SparrowToRiss, CScoreSAT, YalSAT and Score$_2$SAT. Especially, *SelectNTS* solves 150 instances on average, and YalSAT solves 96 (out of 150) instances on average, and Score$_2$SAT solves 50 (out of 150) instances on average, and while SparrowToRiss and CScoreSAT have difficulty in solving these huge random 5-SAT instance classes. In sum, this experiment further confirms the efficiency of *SelectNTS* for solving the general uniform medium and huge random 5-SAT problems.

**Results for Uniform Random 7-SAT**

In Table 11, we show our experimental results on the uniform medium and huge 7-SAT instances.

Table 11 shows that *SelectNTS* is competitive with its competitors for these medium 5-SAT instances. Specifically, *SelectNTS* obtains the best performance for 2 medium instance classes. Score$_2$SAT reaches the best performance for 2 medium instance classes. SparrowToRiss gives the best performance for only one medium instance class. However, the three other algorithms ProbSAT, YalSAT and CScoreSAT performs worse than *SelectNTS* on these uniform medium random 7-SAT instance classes. These results demonstrate that our *SelectNTS* algorithm is quite competitive for solving this medium random 7-SAT problems.

As reported in Table 11, the experimental results show *SelectNTS* dramatically outperforms its competitors. Compared to the competitors whose performance descends steddply as the instance ratio increases, *SelectNTS* shows good scalability. For example, for the huge 7-SAT instances with $r$=66, 67, all competitors fail in solving all instances, while *SelectNTS* solves 50 and 18 huge 7-SAT instances with $r$=66, 67 on average respectively, which confirms the good performance of *SelectNTS* on these huge 7-SAT instances.

**Results for SAT Competition 2018**

Table 12 presents the experimental results of our *SelectNTS* algorithm and SparrowToRiss, CScoreSAT, Score$_2$SAT, YalSAT and ProbSAT on all HRS instances and uniform random $k$-SAT instances with long clauses from SAT Competition 2018 [55]. *SelectNTS* gives the best performance for all random SAT instances except for the HRS instances with $r$=4.3, and especially it solves more uniform random $k$-SAT instances with $k$>3 than all competitors. For the HRS instances with $r$=4.3, *SelectNTS* solves as many instances as Score$_2$SAT and YalSAT but the PAR 2 is a little more than Score$_2$SAT and YalSAT's. Overall, *SelectNTS* solves 209 instances on average, and SparrowToRiss solves 192 instances on average, and Score$_2$SAT solves 137 instances on average, and ProbSAT solves 116 instances on average, and YalSAT solves 111 instances on average, and CScoreSAT solves 106 instances on average. *SelectNTS* significantly outperforms SparrowToRiss on all random SAT instances. SparrowToRiss is the first place on the random SAT track of SAT Competition 2018, thus it is challenging to improve such performance over SparrowToRiss, indicating that *SelectNTS* algorithm achieves the state-of-the-art performance on random SAT instances.

**Summary for HRS and uniform random $k$-SAT with $k$>3**

According to Table5-Table 12, the experimental results show that *SelectNTS* consistently outperforms CScoreSAT, ProbSAT, YalSAT, Score$_2$SAT and SparrowToRiss on solving HRS instances with various ratios and sizes except for the HRS instacnes with $r$=4.3, and is quite competitive for solving uniform random $k$-SAT with long clauses, i.e., the performance of *SelectNTS* in solving HRS instances is better than of that in solving uniform random $k$-SAT with $k$>3. Based on the case study in Section 5, we conjecture that for random SAT problems, if the maximum value of *cNTS* and *vNTS* on *SelectNTS* is less than or equal to **0.5** times as large as on ProbSAT within $10^5$ steps respectively, *SelectNTS* is more effective to solve these problems than other random SAT instances which the maximum value of *cNTS* and *vNTS* on *SelectNTS* is more than **0.5** times as large as on ProbSAT within $10^5$ steps respectively.

## 7. Conclusions and Future Works

In this work, we presented an enhanced probability selecting based on local search method for the well-known HRS problem and uniform random $k$-SAT problem with long clauses. This work has opened up a new direction for effective SLS algorithms. The first enhancement improves the probability selecting approach of the original ProbSAT [6] by using a new and global clause weighting scheme called *cNTS* to distinguish unsatisfied clauses and adopting the biased random walk to prefer satisfying several unsatisfied clauses hard to keep satisfied. The second enhancement concerns the variation of CC strategy, which is both more powerful than the probability selecting algorithm, and utilizes a new and global variable weighting scheme called *vNTS* to distinguish variables and then proposes a linear function named $S_v$ which combines *vNTS* and *Score*, to avoid selecting the same variable in consecutive steps. As the variation of CC strategy, $S_v$ is very simple compared to the CC strategy.

The enhanced probability selecting based on local search method is called *SelectNTS*, whose effectiveness has been demonstrated on random SAT problems from the SAT Competitions in 2017 and 2018, and on randomly generated HRS and uniform $k$-SAT with long clauses problems. The results show that *SelectNTS* outperforms state-of-the-art SLS solvers and the state-of-the-art hybrid solver in most cases. Moreover, *SelectNTS* can effectively solve both uniform random $k$-SAT problems and HRS problems.

As future work, a significant research issue is to improve SLS algorithms for structured problems, constrained satisfaction problems, and graph search problems, by using the new heuristics.

## Acknowledgments

This work is partially supported by the National Natural Science Foundation of China (Grant No. 61673320) and the Fundamental Research Funds for the Central Universities (Grant No. 2682017ZT12, 2682016 CX119, 2682019ZT16, 2682020CX5). The authors would like to thank Tom´aˇs Balyo for providing the generator.